\documentclass[conference]{IEEEtran}
\IEEEoverridecommandlockouts
\usepackage{cite}
\usepackage{amsmath,amssymb,amsfonts}
\usepackage{booktabs}
\usepackage[table]{xcolor}
\usepackage{makecell}
\usepackage{pifont}
\usepackage[caption=false,font=footnotesize]{subfig}
\usepackage{url}
\usepackage{graphicx}
\newcommand{\yesmark}{\textcolor{green!50!black}{\ding{51}}}
\newcommand{\nomark}{\textcolor{red!75!black}{\ding{55}}}
\def\BibTeX{{\rm B\kern-.05em{\sc i\kern-.025em b}\kern-.08em
    T\kern-.1667em\lower.7ex\hbox{E}\kern-.125emX}}
\begin{document}
\title{Droneulator: A Portable UAV Simulator for Agricultural Workflows with RotorPy and Godot 4
\thanks{This work was supported by the Engineering and Physical Sciences Research Council, UK Projects GAIA (Ref: EP/Y003438/1) and AgriFoRwArdS (Ref: EP/S023917/1). Corresponding author: 25105508@students.lincoln.ac.uk}
}

\author{
\IEEEauthorblockN{
Jacob Swindell\IEEEauthorrefmark{1}, 
Michael Lowen\IEEEauthorrefmark{1}, 
Marija Popovi\'{c}\IEEEauthorrefmark{2}, and 
Riccardo Polvara\IEEEauthorrefmark{1}
}
\IEEEauthorblockA{\IEEEauthorrefmark{1}\textit{Lincoln Centre for Autonomous Systems (L-CAS)}, \textit{University of Lincoln}, Lincoln, UK.}

\IEEEauthorblockA{\IEEEauthorrefmark{2}\textit{Faculty of Aerospace Engineering, MAVLab}, \textit{Delft University of Technology (TU Delft)}, Delft, Netherlands.}
}

\IEEEaftertitletext{%
  \vspace{-0.4em}
  \begin{center}
  \includegraphics[width=0.9\textwidth]{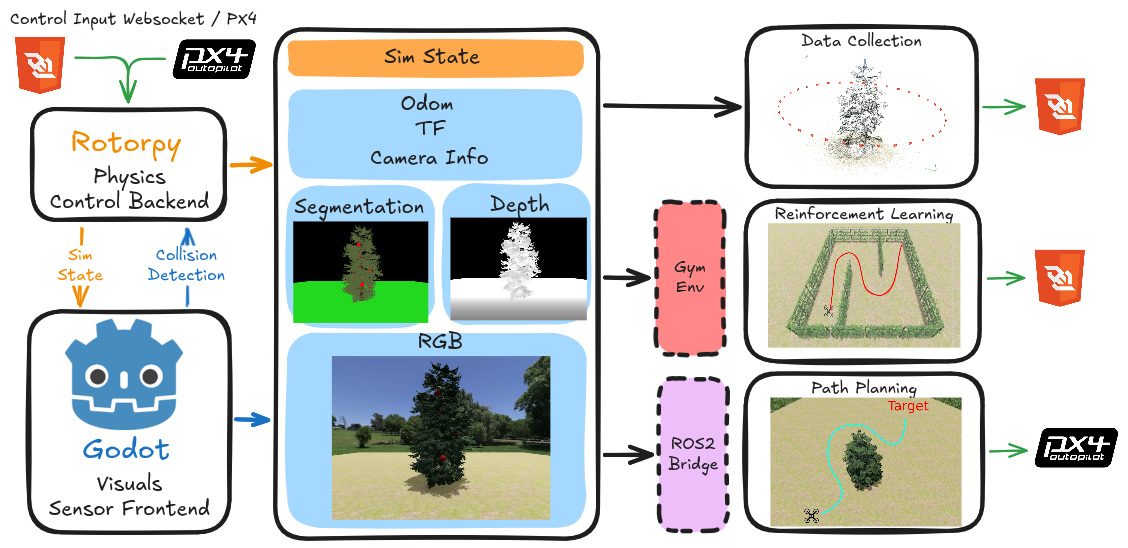}\\
  \vspace{0.25em}
  \refstepcounter{figure}\footnotesize \textbf{Fig.~\thefigure.} System architecture overview of Droneulator. RotorPy provides multirotor dynamics, Godot 4 provides rendering and sensors, Zenoh exposes ROS~2-compatible sensing, and commands arrive through either PX4 SITL or a lightweight WebSocket path, enabling a portable workflow that supports both PX4-based control and lightweight scripted experiments. \label{fig:architecture}
  \end{center}
  \vspace{0.7em}
}

\maketitle

\begin{abstract}
Agricultural UAV research requires simulators that integrate realistic 3D scenes, high-fidelity vehicle dynamics, and robotics middleware, while remaining practical to deploy across heterogeneous development machines. We present Droneulator, a portable UAV simulator architecture that combines RotorPy~\cite{folk2023rotorpy} for multirotor dynamics with Godot 4~\cite{godot_engine} for rendering and sensor generation. Droneulator exposes both PX4-based control~\cite{px4_docs} and a lightweight WebSocket command path, and publishes synchronised visual and state streams through a Zenoh-based ROS~2-compatible pipeline~\cite{eclipse_zenoh,zenoh_bridge_dds}. This integration enables a single stack to support inspection-oriented data capture, ROS~2/PX4 local planning, and reinforcement learning experiments without modifying the simulator infrastructure. We present quantified validation of the current system across three agricultural UAV workflows: tree-scale image collection for 3D reconstruction with COLMAP~\cite{schoenberger2016sfm}, local planning around canopy obstacles using EGO-Planner~\cite{zhou2021ego}, and closed-loop reinforcement learning through a custom Gymnasium environment. In the reported setup, the results show that the simulator can sustain low-latency sensing, support reconstruction-oriented data collection under varying capture density, execute collision-free local planning around canopy obstacles, and support stable depth-sensing-based policy training for obstacle-aware navigation. Together, these results show the potential of Droneulator for agricultural UAV inspection, planning, and learning within one deployable stack.
\end{abstract}

\begin{IEEEkeywords}
Systems simulation, Drones, Autonomous aerial vehicles, Agricultural robots
\end{IEEEkeywords}

\section{Introduction}

Agricultural UAV research must bridge orchard-like 3D scenes, flight dynamics suitable for controller development, and interfaces that plug into existing tooling for perception, planning, and learning. These requirements are especially relevant for canopy inspection, local navigation around vegetation, and data collection for digital-twin generation, where field-testing is expensive and operationally constrained.

Recent aerial robotics simulator surveys emphasise tradeoffs among rendering fidelity, vehicle dynamics, integration readiness, and workflow usability~\cite{dimmig2025survey}. A recent PX4 community report likewise highlights multi-OS support, ROS~2 connectivity, and ease of installation as recurring priorities~\cite{mcguire2025px4survey}. For the agricultural workflows considered here, the practical difficulty is not finding simulators that satisfy one of these requirements in isolation, but finding one stack that combines deployable packaging, UAV-oriented dynamics, visual sensing, ROS~2/PX4 integration, and support for both scripted and learning-based experiments. Droneulator\footnote{Project repository: \url{https://github.com/JakeSwin/Droneulator}. The codebase will be made publicly available upon acceptance.} is designed around that integration gap while avoiding a native ROS~2 dependency on the simulator host. We show that this integrated architecture is sufficient to support three representative agricultural UAV workflows without changing the simulator infrastructure. 

\begin{figure*}[!t]
    \centering
    \subfloat[Main menu]{\includegraphics[width=0.24\textwidth]{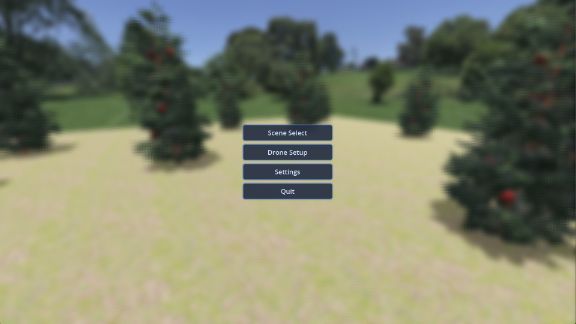}\label{fig:main_menu}}
    \hfill
    \subfloat[Level select]{\includegraphics[width=0.24\textwidth]{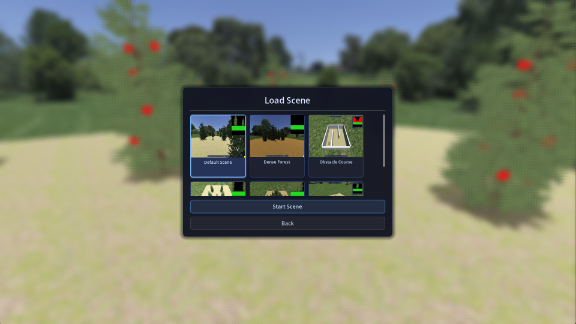}\label{fig:load_scene}}
    \hfill
    \subfloat[Default scene]{\includegraphics[width=0.24\textwidth]{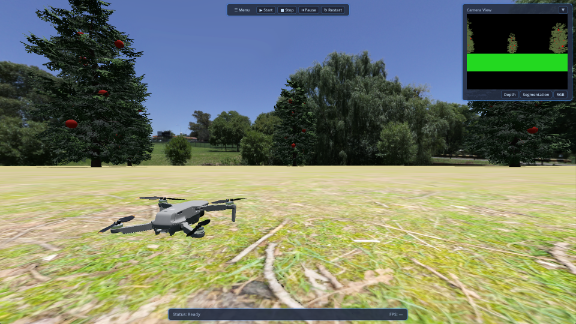}\label{fig:default_scene}}
    \hfill
    \subfloat[Tree inspection scene]{\includegraphics[width=0.24\textwidth]{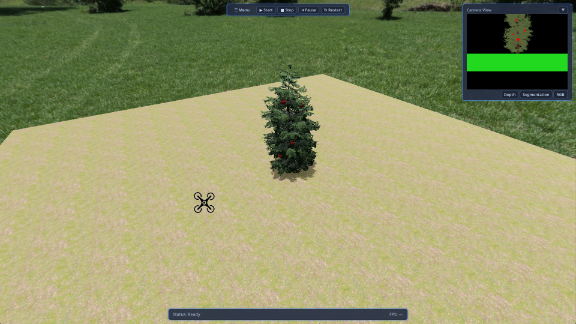}\label{fig:inspect_scene}}
    \caption{Droneulator user interface and representative agricultural UAV scenes used for inspection-oriented experiments.}
    \label{fig:sim_screenshots}
\end{figure*}

Existing simulators cover different parts of this design space. As summarised in Table~\ref{tab:sim_comparison}, broad robotics platforms such as Gazebo~\cite{koenig2004design} and Webots~\cite{michel2004cyberbotics} provide mature ecosystems and middleware connectivity, but not the same UAV-focused workflow coverage across inspection, planning, and learning. Visually rich platforms such as Isaac Sim~\cite{NVIDIA_Isaac_Sim} and AirSim~\cite{shah2017airsim} provide strong rendering and sensing, but they do not offer the same combination of lightweight deployment and first-party integration paths targeted here. More specialised UAV or RL-oriented simulators, including Flightmare~\cite{song2021flightmare}, RotorS~\cite{furrer2016rotors}, and gym-pybullet-drones~\cite{panerati2021learning}, support parts of this stack but remain narrower in middleware, sensing, or deployment scope. Table~\ref{tab:sim_comparison} therefore serves as a coarse positioning context; the paper's main argument comes from the quantified workflow results that follow.


\begin{table}[!t]
\caption{Feature comparison of representative UAV simulators. Marks denote direct, first-party support in mainstream releases.~\cite{dimmig2025survey,mcguire2025px4survey}}
\label{tab:sim_comparison}
\centering
\resizebox{\columnwidth}{!}{%
\begin{tabular}{lccccccc}
\toprule
\textbf{Simulator} & \makecell{\textbf{Multi-OS}\\\textbf{avail.}} & \makecell{\textbf{UAV}\\\textbf{dynamics}} & \makecell{\textbf{ROS}\\\textbf{2}} & \makecell{\textbf{PX4}} & \makecell{\textbf{Sensor}\\\textbf{suite}} & \makecell{\textbf{RL}\\\textbf{workflow}} & \makecell{\textbf{Active}\\\textbf{upstream}} \\
\midrule
Gazebo~\cite{koenig2004design} & \yesmark & \yesmark & \yesmark & \yesmark & \yesmark & \nomark & \yesmark \\
Isaac Sim~\cite{NVIDIA_Isaac_Sim} & \yesmark & \yesmark & \yesmark & \nomark & \yesmark & \yesmark & \yesmark \\
AirSim~\cite{shah2017airsim} & \yesmark & \yesmark & \nomark & \yesmark & \yesmark & \yesmark & \nomark \\
Flightmare~\cite{song2021flightmare} & \nomark & \yesmark & \nomark & \nomark & \yesmark & \yesmark & \nomark \\
RotorS~\cite{furrer2016rotors} & \nomark & \yesmark & \nomark & \nomark & \nomark & \nomark & \nomark \\
Webots~\cite{michel2004cyberbotics} & \yesmark & \nomark & \yesmark & \nomark & \yesmark & \nomark & \yesmark \\
gym-pybullet-drones~\cite{panerati2021learning} & \yesmark & \yesmark & \nomark & \nomark & \nomark & \yesmark & \yesmark \\
\textbf{Droneulator} & \yesmark & \yesmark & \yesmark & \yesmark & \yesmark & \yesmark & \yesmark \\
\bottomrule
\end{tabular}
}
\end{table}


In Table~\ref{tab:sim_comparison}, we evaluate representative UAV simulators across key dimensions. `Multi-OS avail.' indicates documented, first-party support on multiple desktop operating systems. `UAV dynamics' refers to dedicated, high-fidelity multirotor flight dynamics rather than generic rigid-body kinematics. `ROS 2' and `PX4' denote out-of-the-box integration with the ROS 2 ecosystem and PX4 Autopilot (e.g., via SITL). `Sensor suite' highlights comprehensive rendering support for perception, including RGB, depth, and semantic segmentation. `RL workflow' designates documented, closed-loop training environments for reinforcement learning, distinct from generic scriptability. Finally, `Active upstream' signifies the project is actively maintained and updated by its developers.

Our contribution is a deployable simulator architecture that unifies scene rendering, multirotor dynamics, ROS 2-compatible sensing, and dual control paths within one portable stack. Concretely, this paper makes three contributions. First, we present a deployable simulator architecture that combines Godot 4 scene rendering, RotorPy multirotor dynamics, and dual control paths within one workflow. Second, we provide a Zenoh-based sensing pipeline that exposes synchronised visual and state data in ROS~2-compatible form, including RGB, depth, semantic segmentation, odometry, camera information, and TF~\cite{eclipse_zenoh,zenoh_bridge_dds}. Third, we report quantified validation across three agricultural UAV workflows in tree-rich scenes: inspection-oriented data collection, local planning, and reinforcement learning.

\section{System Architecture \& Design}

Droneulator separates dynamics and rendering so that UAV physics, scene authoring, and downstream robotics workflows can evolve independently within a portable deployment-oriented setup. Figure~\ref{fig:architecture} summarises how the simulator connects Godot-rendered sensing, RotorPy dynamics, and external clients for data collection, planning, and learning.

\subsection{Dynamics Engine}


RotorPy~\cite{folk2023rotorpy} provides the multirotor dynamics backend, including aerodynamic effects and controller-oriented simulation. It was chosen because it is a modern, actively maintained, and intuitive multirotor dynamics engine that easily extends to complex tasks. To preserve portability, the RotorPy loop is packaged as a standalone executable using PyInstaller and launched by Godot as a child process. Because Godot manages the binary directly, startup and shutdown remain synchronised with the visual frontend without requiring an external Python environment on the host machine.

Godot exchanges pose, attitude, and command data with the dynamics process over WebSocket, beginning with a handshake that identifies the simulator client. Because RotorPy uses a right-handed $Z$-up frame while Godot uses a right-handed $Y$-up, $-Z$-forward frame, the state stream is converted online before being rendered. This decoupled architecture makes the flight model reusable across control and learning experiments while keeping scene and sensor development flexible.

\subsection{Sensor Pipeline}
\label{sec:sensor_pipeline}

Godot provides environment creation and sensor generation for scripted outdoor scenes, including tree-rich agricultural settings. A three-camera rig supplies RGB images compressed with JPEG, semantic-segmentation images transmitted as lossless PNG, and metric depth reconstructed from the Z-buffer and transmitted raw. The segmentation stream is produced from a duplicated rendering layer whose meshes reuse the original geometry while replacing materials with flat class colours, keeping the implementation lightweight while preserving pixel-level labels for downstream perception tasks.

Together with RotorPy odometry, these streams are published over Zenoh in ROS~2-compatible CDR format~\cite{eclipse_zenoh,zenoh_bridge_dds}. The simulator also publishes TF and camera information, enabling perception and planning stacks to consume synchronised topics at 30~Hz. This is particularly useful for agri-UAV workloads in which image streams, geometry, and pose must remain temporally aligned for reconstruction, navigation, or policy learning.

\begin{table}[htbp]
    \centering
    \caption{End-to-end latency for the Zenoh-ROS~2 bridge over 60~s at 30~Hz ($n=1800$ samples per stream).}
    \label{tab:latency_metrics}
    \begin{tabular}{l c c c}
        \toprule
        \textbf{Sensor Stream} & \textbf{Payload Type} & \textbf{Mean (ms)} & \textbf{Std. Dev. (ms)} \\
        \midrule
        Odometry & State Struct & 0.45 & $\pm$0.12 \\
        RGB Camera & Compressed (JPEG) & 5.65 & $\pm$0.29 \\
        Depth Camera & Raw (Float32) & 17.06 & $\pm$1.23 \\
        \bottomrule
    \end{tabular}
\end{table}

To quantify middleware efficiency, we measured end-to-end sensing latency on the local host machine over a 60-second execution window. Table~\ref{tab:latency_metrics} shows sub-millisecond mean odometry latency (0.45~ms) and 17.06~ms mean latency for raw depth, which remained below the 33.3~ms budget associated with a 30~Hz update rate in the reported setup. The higher depth latency is consistent with the larger raw payload, but it still leaves headroom for the planning and reinforcement learning loops used elsewhere in the paper. These measurements show that the bridge is sufficient for the reported workflows. 

\begin{figure}[!ht]
    \centerline{\includegraphics[width=0.78\columnwidth]{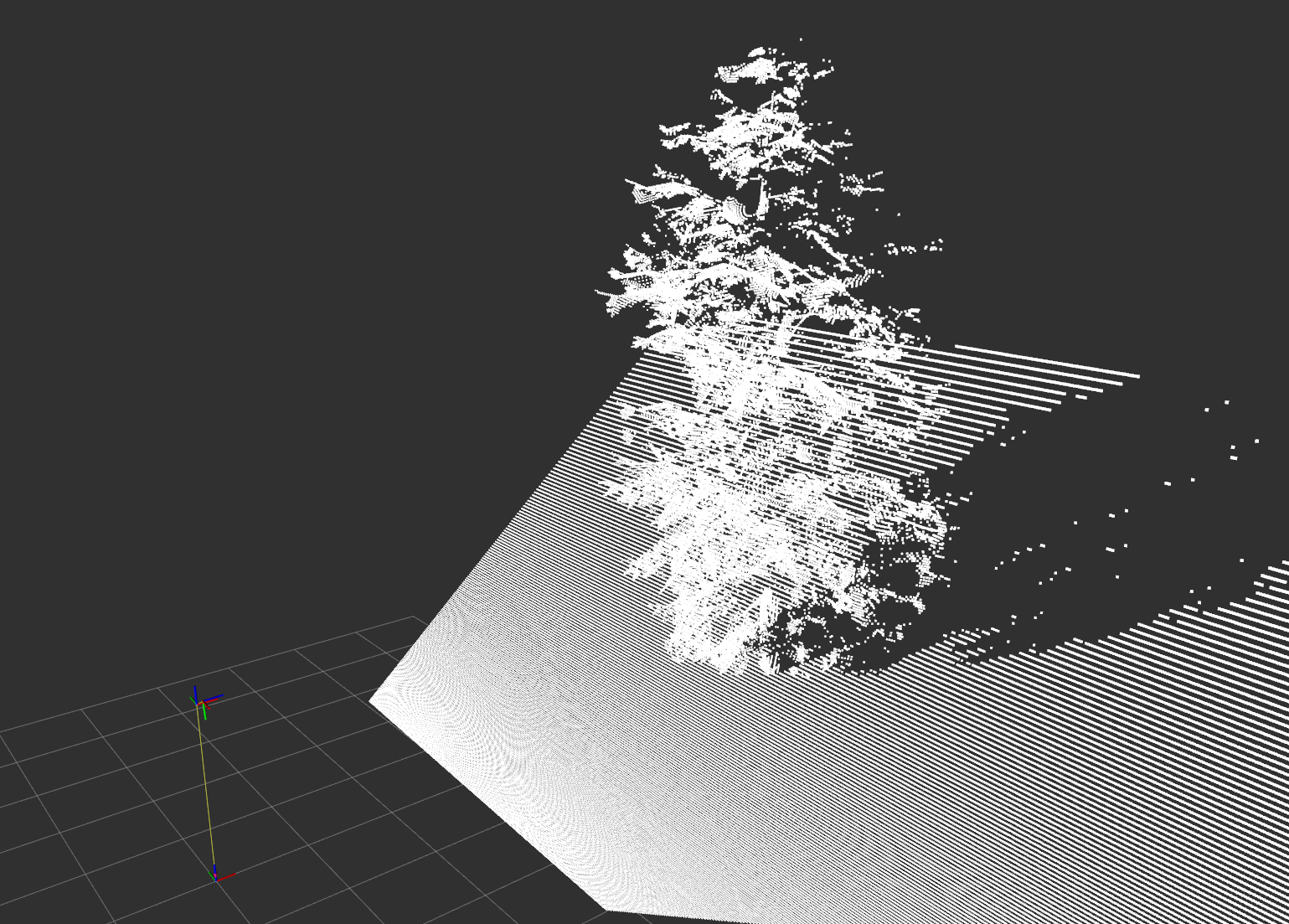}}
    \caption{Orchard-style tree scene visualised in RViz from the simulator's depth output.}
    \label{fig:rviz}
\end{figure}

\subsection{Control Interface}

The control stack is intentionally modular. Droneulator supports a PX4 bridge for software-in-the-loop control~\cite{px4_docs}, allowing operation through standard tools such as QGroundControl, and it also exposes a lightweight WebSocket interface for high-level commands.

This interface operates in two modes. The first is a position-and-yaw mode executed via a minimum-snap trajectory~\cite{mellinger2011minimum} and an internal SE(3) controller~\cite{lee2010geometric}, which is sufficient for move-to-point data-collection workflows. Minimising snap is highly critical for multirotors because it guarantees smooth transitions in motor thrusts and body torques, thereby reducing mechanical vibrations and ensuring the controller can accurately track the intended path.

The second control mode uses a continuous velocity-control mode for reinforcement learning agents, where a unicycle-style command pair is smoothed through slew-rate and low-pass filters before being integrated into a virtual reference setpoint tracked by the SE(3) controller. 

This split supports the paper’s target workflows: PX4 integration enables compatibility with existing autonomy and operator tooling, while the lightweight command interface supports both repeatable scripted experiments and dynamic reinforcement learning training without requiring a full autopilot stack.

\subsection{Packaging and Portability}

Portability is a core design requirement, defined here as the ability to deploy across multiple operating systems while minimising host dependencies by bundling necessary components directly into the application. Accordingly, the Godot frontend and RotorPy backend are packaged as native binaries, and Zenoh decouples the simulator from the ROS~2 environment. This architecture enables both lightweight scripted experiments and full ROS~2/PX4 deployments.

\section{Agricultural Workflows}

The following examples provide quantified systems validation across three agricultural workflows. The reported measurements were collected on an x86 Linux desktop PC (AM5 7700 CPU, RTX 4070 GPU). 


\subsection{Automated Data Collection}

To demonstrate Droneulator's utility for agricultural inspection and data collection, we developed a pipeline for 3D point cloud generation using COLMAP~\cite{schoenberger2016sfm}. A test application computes a circular image-capture trajectory around a tree, sends the resulting waypoints over the WebSocket interface, and records synchronised RGB images and poses as the vehicle reaches each target viewpoint. Internally, the simulator executes these move-and-capture commands through the same state and control interfaces exposed to downstream robotics software, which makes the workflow representative of a practical orchard inspection stack.

The capture client uses the published camera information and simulation-state topics to determine when each waypoint has been reached and when an image should be recorded. Because the simulator publishes synchronised ground-truth odometry alongside the visual stream, we supplied these poses to COLMAP as priors, mirroring inspection settings in which accurate localisation is available from RTK-GPS, visual-inertial odometry, or fused state estimation. In practice, these priors stabilise the reconstruction pipeline and reduce failure modes associated with poor initialisation or scale drift. Across the reported task, all three image collection runs used the same tree scene, circular capture policy, image resolution, and COLMAP pipeline. Only the number of sampled viewpoints changed.

\begin{figure}[!ht]
    \centering
    \subfloat[18 images]{\includegraphics[trim={4cm 4cm 4cm 4cm}, clip, width=0.32\columnwidth]{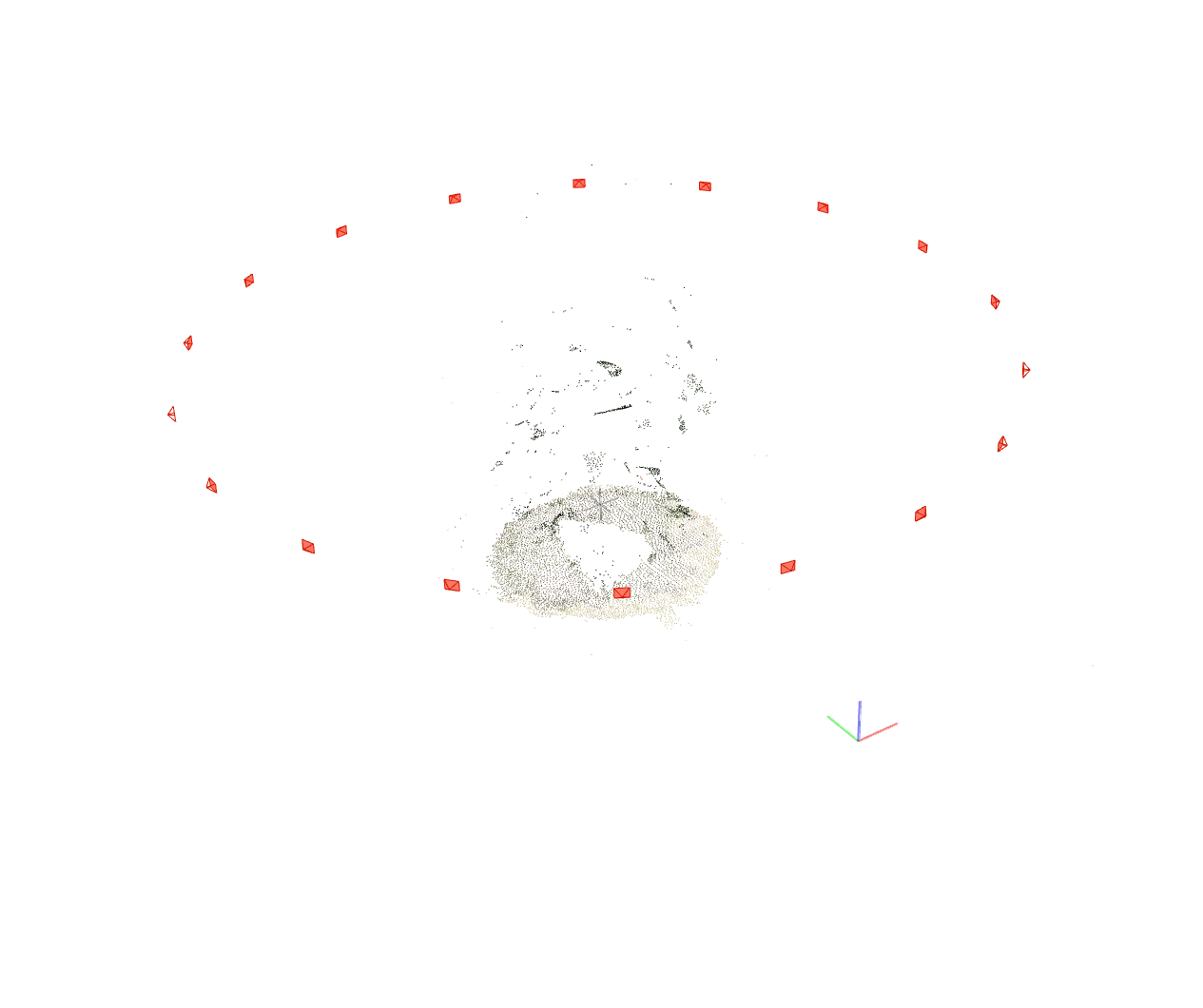}}\label{fig:colmap_18}
    \hfill
    \subfloat[36 images]{\includegraphics[trim={4cm 4cm 4cm 4cm}, clip, width=0.32\columnwidth]{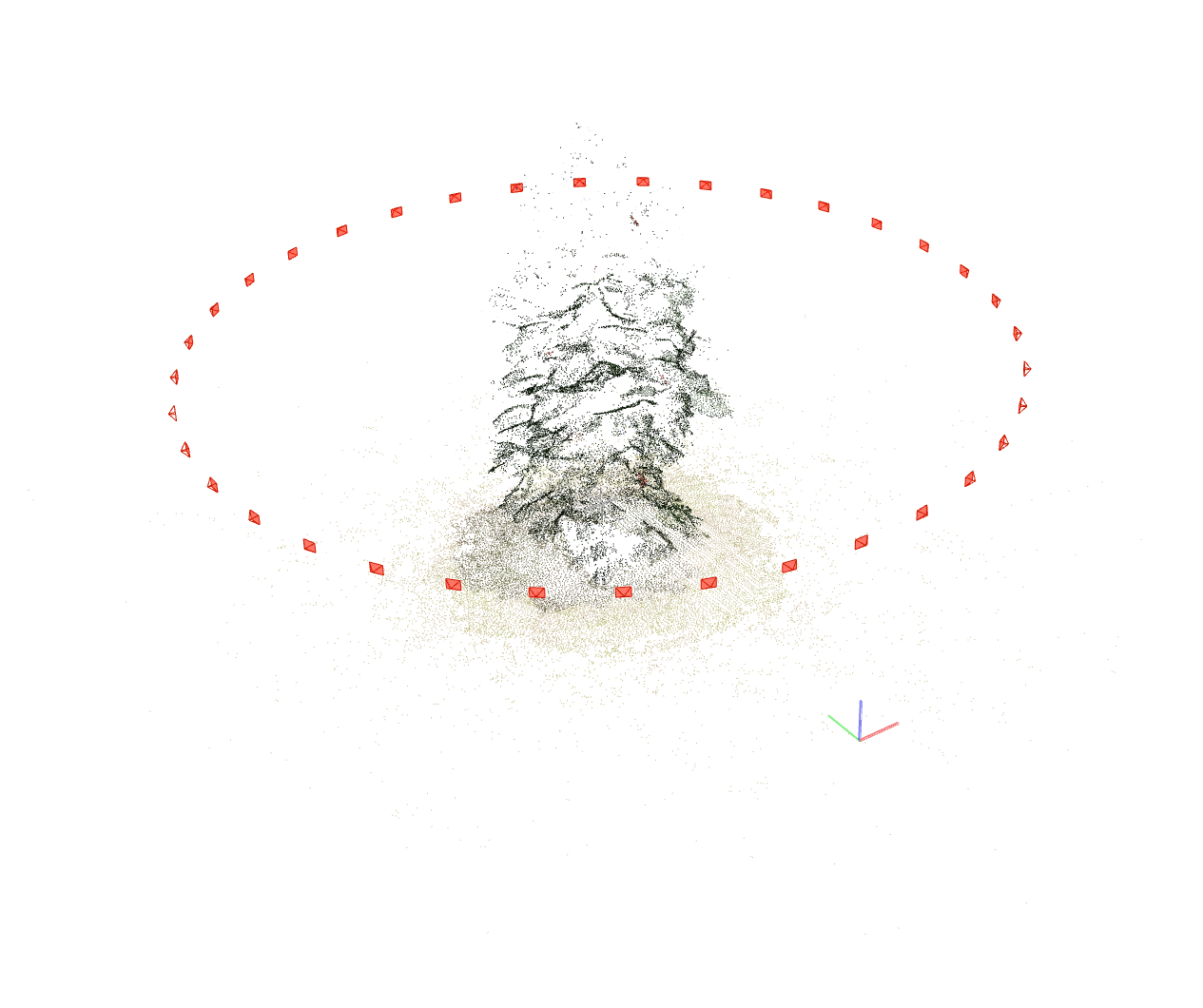}}\label{fig:colmap_36}
    \hfill
    \subfloat[54 images]{\includegraphics[trim={4cm 4cm 4cm 4cm}, clip, width=0.32\columnwidth]{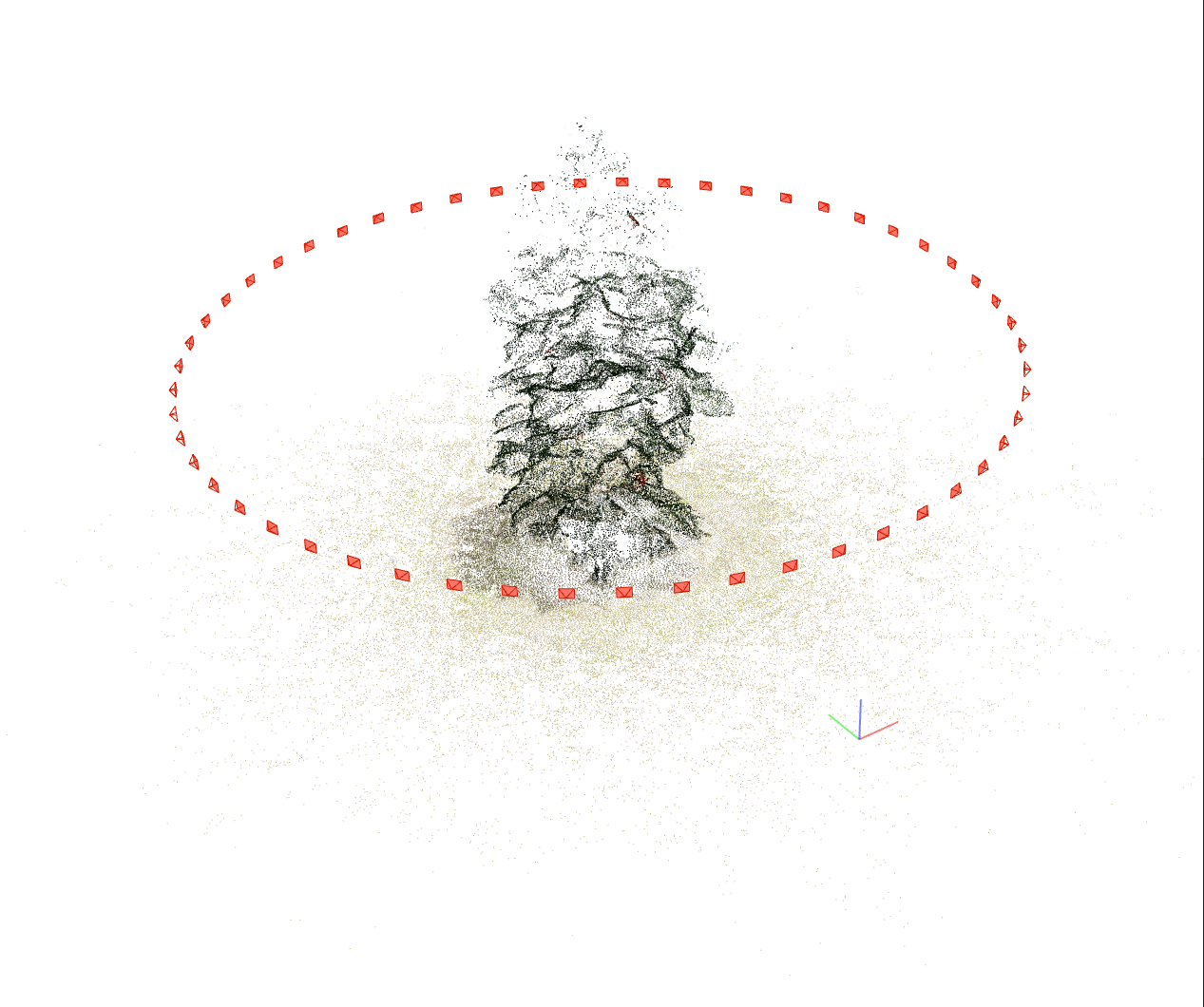}}\label{fig:colmap_54}
    
    \caption{Dense COLMAP reconstructions from Droneulator-captured RGB images at increasing viewpoint densities, showing improved canopy completeness and point density with denser capture.}
    \label{fig:dense_reconstruct}
\end{figure}

\begin{table}[htbp]
    \centering
    \caption{3D reconstruction quality and computational cost across varying data-collection densities under a fixed scene and COLMAP configuration.}
    \label{tab:colmap_results}
    \resizebox{\columnwidth}{!}{%
    \begin{tabular}{c c c c c c}
        \toprule
        \textbf{Images} & \textbf{Sparse Pts} & \textbf{Dense Pts} & \textbf{Mean Track} & \textbf{Reconstruction Time (min)} & \textbf{Reproj Error (px)} \\
        \midrule
        18 & 149 & 6,518 & 3.09 & $\sim$3.9 & 0.79 \\
        36 & 1,156 & 34,795 & 3.72 & $\sim$8.4 & 0.74 \\
        54 & 2,760 & 79,073 & 4.15 & $\sim$11.8 & 0.72 \\
        \bottomrule
    \end{tabular}%
    }
\end{table}

Figure~\ref{fig:dense_reconstruct} and Table~\ref{tab:colmap_results} demonstrate how varying waypoint density impacts reconstruction. Increasing sampling from 18 to 54 images raises the mean track length to 4.15 and yields 79,073 dense points with a 0.72-pixel reprojection error. However, increasing from 36 to 54 images reveals diminishing returns. The dense-point count doubles, yet reprojection error improves only slightly. While denser sampling improves canopy completeness in the reported tree scene, it ultimately increases execution cost without proportional accuracy gains once baseline coverage is met. Additionally, this dense 54-image configuration exposed the limits of RotorPy's SE(3) controller, with the tightly packed trajectories causing flight instability and altitude drops during waypoint transitions.

\subsection{Local Planning with ROS~2 and PX4}

To demonstrate compatibility with external autonomy stacks, we integrated EGO-Planner~\cite{zhou2021ego}, a widely used open-source planner, through ROS~2 while using PX4 for low-level control. Using the sensing pipeline described in Section~\ref{sec:sensor_pipeline}, the simulator exposes the image, odometry, TF, and camera-information streams required by the planner.

\begin{figure}[!ht]
    \centerline{\includegraphics[width=0.95\columnwidth]{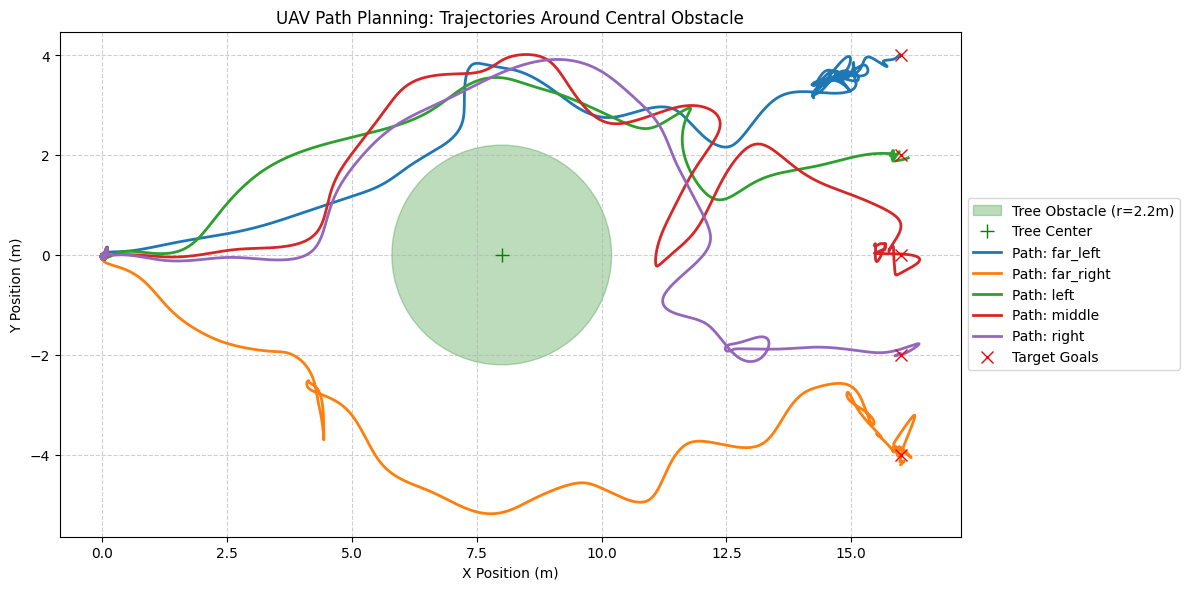}}
    \caption{Executed UAV odometry from five EGO-Planner runs reaching distinct goals while avoiding a central tree obstacle.}
    \label{fig:ego_planner_traj}
\end{figure}

This use case is executed by starting PX4 SITL, the simulator, and the Zenoh ROS~2 bridge in a ROS~2 Jazzy environment. Because EGO-Planner does not output PX4-native commands directly, we use Micro XRCE-DDS Agent to forward the planner's ROS~2 command stream into the PX4 control path. After takeoff, the vehicle can be assigned goal positions via RViz or the ROS~2 command line interface and then navigate around nearby tree obstacles. As shown in Fig.~\ref{fig:ego_planner_traj}, the five executed trajectories bend around the central tree before converging to their respective goal regions. This multi-goal scenario highlights a complete sensing-and-control chain spanning depth rendering, TF publication, ROS~2 planning, PX4 control, and RotorPy execution, demonstrating Droneulator's ability to support complex, modern ROS~2 autonomy stacks.

For the metrics reported below, goal convergence is the Euclidean distance between the final executed UAV position and the commanded goal, while obstacle clearance is the minimum Euclidean distance between the executed trajectory and the tree boundary over the full run.

\begin{table}[htbp]
    \centering
    \caption{Trajectory statistics for multi-goal local planning using EGO-Planner.}
    \label{tab:ego_planner_metrics}
    \resizebox{\columnwidth}{!}{%
    \begin{tabular}{l c c c}
        \toprule
        \textbf{Target Region} & \textbf{Flight Time (s)} & \textbf{Goal Convergence (m)} & \textbf{Obstacle Clearance (m)} \\
        \midrule
        Far Left & 59.80 & 0.008 & 0.286 \\
        Left & 23.76 & 0.071 & 1.017 \\
        Middle & 35.18 & 0.026 & 0.883 \\
        Right & 34.35 & 0.005 & 1.089 \\
        Far Right & 39.43 & 0.002 & 2.182 \\
        \bottomrule
    \end{tabular}%
    }
\end{table} 

As detailed in Table~\ref{tab:ego_planner_metrics}, the multi-goal trajectory tests routed the UAV around the central canopy without collision, maintaining a minimum observed clearance of 0.28~m. Flight times ranged from 23.76~s to 59.80~s, indicating obstacle-dependent trajectories rather than uniform paths, while terminal accuracy remained sub-decimeter across all runs with an average goal-convergence error of 0.02~m. The spread across target regions is also informative: the far-left case produced both the longest flight and the tightest clearance, whereas the far-right route preserved over 2~m of separation, showing that the same planner-controller stack adapts to different local free-space geometries around a shared obstacle. In the reported setup, the ROS~2/PX4 pipeline therefore achieved successful local obstacle avoidance across multiple goals while preserving positive clearance in canopy-like clutter. 

\subsection{Reinforcement Learning}

To demonstrate learning-based integration, we constructed a custom Gymnasium environment for closed-loop training. The environment communicates with the simulator via a WebSocket client for UAV state and control, alongside Zenoh subscribers for image sensing, ensuring that each step provides an up-to-date observation vector to the RL agent.

We formulated an obstacle-aware goal-reaching task within a simulated agricultural vineyard environment and trained a baseline Soft Actor-Critic (SAC) agent over 50,000 steps. Instead of relying purely on physical state variables, the agent perceives the environment through simulated sensor outputs, providing a vision-in-the-loop training interface for the reported task. 

This environment setup is particularly advantageous for agricultural robotics. Navigating vineyard rows presents a highly constrained challenge where reliable depth perception and precise flight control are critical to avoid damaging crops. By coupling Godot’s high-fidelity depth rendering of complex canopy structures with RotorPy’s realistic multirotor dynamics, Droneulator accurately captures the physical and perceptual realities of flying in tight proximity to foliage.

The observation space is defined as a continuous vector $o_t \in \mathbb{R}^{35}$, constructed as:
$$o_t = [D, \theta_{err}, \tilde{v}_{fwd}, \tilde{\omega}_{yaw}]^T$$
where $D \in [0, 1]^{32}$ is a downsampled, normalised 1D depth strip extracted from the centre row of the simulator's rendered depth camera image (clipped at a 10-metre maximum range). The variable $\theta_{err} \in [-1, 1]$ represents the normalised relative yaw angle to the goal, while $\tilde{v}_{fwd}$ and $\tilde{\omega}_{yaw}$ represent the normalised forward velocity and yaw rate, respectively. 

The continuous action space $a_t \in \mathbb{R}^2$ directly commands the UAV's forward velocity and yaw rate:
$$a_t = [v_{fwd}, \omega_{yaw}]^T$$
where $v_{fwd} \in [-2.0, 3.0]$ m/s and $\omega_{yaw} \in [-1.5, 1.5]$ rad/s. Altitude is held constant by the simulator's underlying flight controller, isolating the 2D navigation challenge.

To guide learning, we used a dense reward function augmented by a programmatic A* pathfinding heuristic computed over a discretised grid of the vineyard canopy. The reward function $R_t$ at step $t$ is defined as:
$$R_t = \lambda(d_{t-1} - d_t) - p_{step} + R_{term}$$
Here, $d_t$ is the current A* shortest-path distance to the goal, and $\lambda = 1.0$ scales the progress reward. A step penalty $p_{step} = 0.01$ encourages efficiency. The terminal reward $R_{term}$ is defined by the environment's termination conditions, which rely on Godot's internal collision detection relayed back to the Python process:
$$R_{term} = \begin{cases} 100 & \text{if distance to goal} < 1.5 \text{ m} \\ -30 & \text{if collision detected} \\ 0 & \text{otherwise} \end{cases}$$

The training curves reported below correspond to a single 50,000-step SAC run. The final-trajectory statistics are computed from 10 inference rollouts of the trained policy.

\begin{figure}[!ht]
    \centerline{\includegraphics[width=0.95\columnwidth]{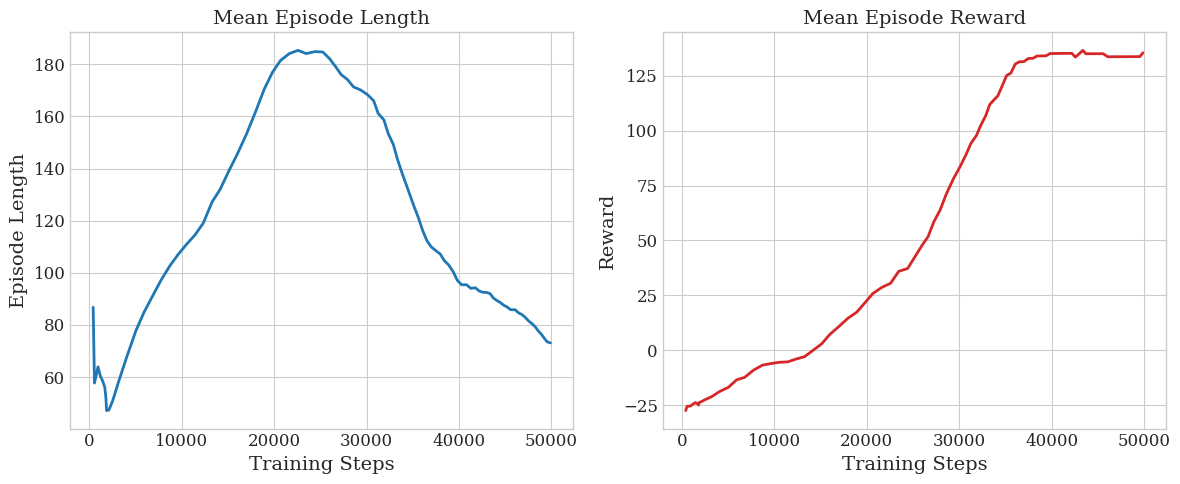}}
    \caption{Training curves from a single 50,000-step SAC run show improving reward and reduced episode length after initial obstacle-avoidance learning, consistent with convergence toward efficient goal-reaching behaviour.}
    \label{fig:rl_training_results}
\end{figure}

Figure \ref{fig:rl_training_results} shows two training phases. Mean episode length first increases as the agent learns basic survival and obstacle avoidance, then decreases once the policy begins reaching the goal more efficiently. Mean episode reward rises steadily and then stabilises, consistent with convergence toward an efficient collision-free policy.

\begin{figure}[!ht]
    \centerline{\includegraphics[width=0.95\columnwidth]{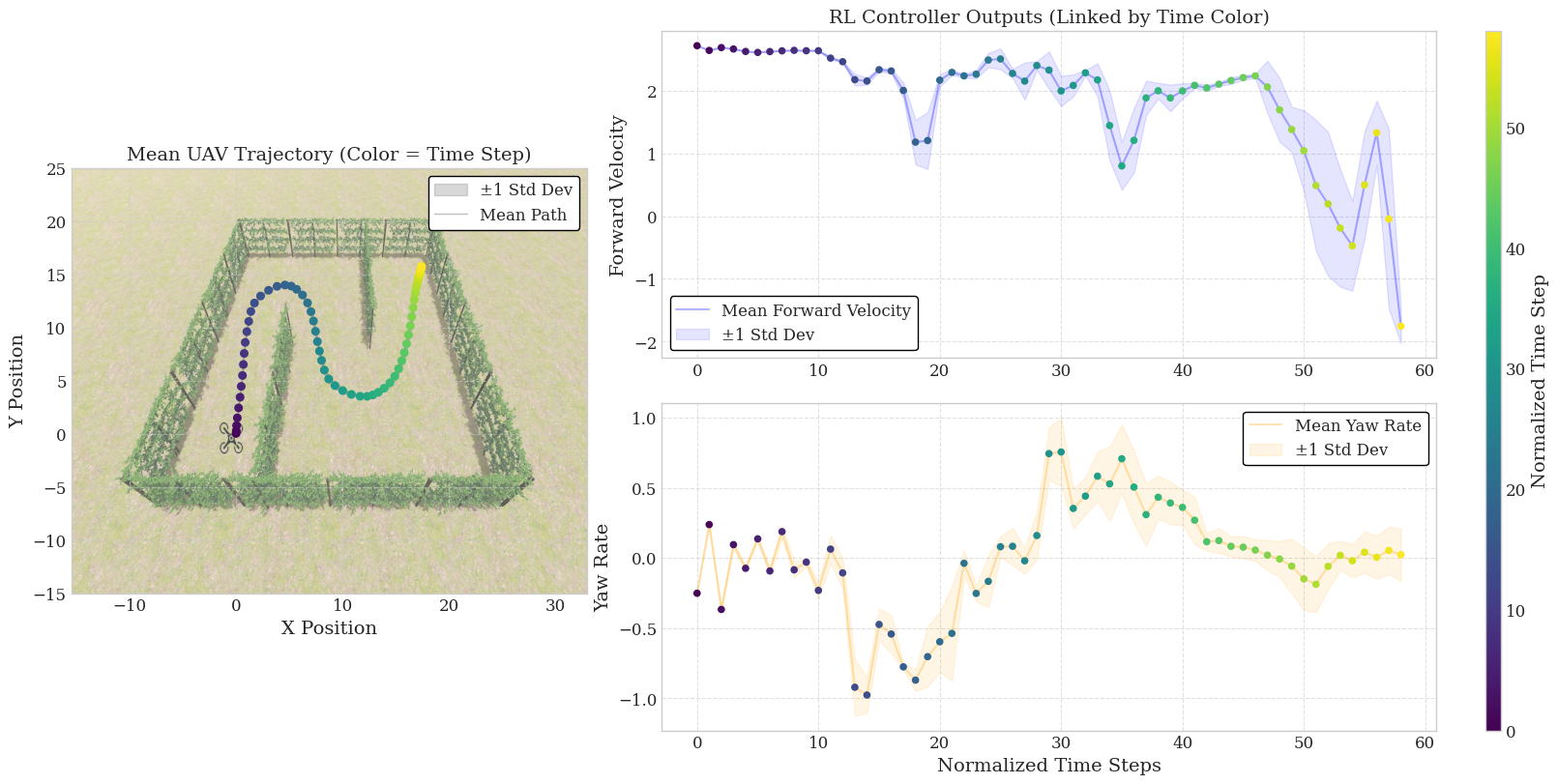}}
    \caption{Spatial and kinematic performance across 10 successful inference rollouts of the trained policy. Left: Multirotor trajectory through the simulated vineyard. Right: Forward velocity and yaw-rate commands, colour-coded to correspond with positions along the spatial path.}
    \label{fig:rl_trajectory_results}
\end{figure}

Figure \ref{fig:rl_trajectory_results} illustrates the spatial and kinematic behaviour of the trained agent. The mean trajectory navigates between the vine canopies while the bounded standard deviations across 10 inference runs highlight stable forward-velocity and yaw-rate outputs. The RL workflow therefore complements the scripted inspection and PX4-coupled planning results: it exercises the lightweight control path, depth sensing, collision handling, and reset logic under repeated interactions rather than single scripted passes. The reported setup produced 10/10 successful evaluation rollouts, demonstrating stable, closed-loop training using depth-sensing in the loop.

Droneulator supports repeated depth-based closed-loop training episodes with reset, sensing, and control in the loop. The observation and action design are intentionally compact. By using a 1D depth profile, relative heading error, and low-dimensional velocity commands, the experiment isolates the simulator's sensing, control, and reset interfaces from the separate challenge of training a large visual encoder. This makes the result appropriate for the paper's purpose: it shows that the simulator can sustain repeated, sensor-driven learning episodes with meaningful canopy interactions, while leaving richer end-to-end vision policies as a natural extension rather than a prerequisite for validation. The RL findings should therefore be read as evidence of stable closed-loop training for the reported task, with limits given by the single-seed training run and 10-rollout evaluation.

\section{Discussion}

Having established Droneulator as a versatile integration layer across inspection, planning, and learning workflows, we identify several key areas for ongoing development to further extend the platform's capabilities.

\paragraph{Sim-to-Real Gap and Environmental Modelling} Presently, the simulator provides a highly controlled environment with ideal aerodynamic conditions and deterministic sensing. Real-world agricultural deployments must contend with dynamic wind disturbances, highly variable lighting, and significant sensor noise—particularly concerning depth-camera degradation when observing thin branches or sparse foliage. Future iterations will introduce configurable wind profiles into the RotorPy backend and statistical noise models into the Godot sensor outputs to better replicate the perceptual and physical disturbances typical of field robotics.

\paragraph{Algorithmic and Controller Boundaries} Our experiments highlighted the operational boundaries of the current control implementation. As observed during the dense 54-image data collection, highly aggressive, tightly-packed spatial trajectories strained the internal SE(3) controller. While this effectively demonstrates the simulator’s physical realism by enforcing kinematic constraints, users testing high-speed or highly dynamic trajectories may need to tune the controller gains or bypass it entirely in favour of direct motor-command inputs. Furthermore, the EGO-Planner demonstration successfully validated canopy-aware local avoidance but did not address global routing across expansive orchard maps, which remains an area for future user-driven integration.

\paragraph{Scalability and Execution Speed} To maintain synchronised physics and visual rendering across separated processes, the current architecture operates at real-time execution speeds within single-agent environments. While this synchronous lockstep is highly effective for prototyping vision-in-the-loop policies and validating single-UAV navigation stacks, expanding the platform to support massive, faster-than-real-time parallel reinforcement learning or multi-agent swarm simulations is a primary objective for future architectural updates.

\section{Conclusion}
Droneulator provides a unified, deployable UAV simulation architecture tailored for agricultural inspection, planning, and learning in complex outdoor environments. By integrating RotorPy, Godot 4, a Zenoh-bridged ROS~2 pipeline, and dual control paths, it effectively collapses the toolchain required for multi-goal data collection, local canopy-aware planning, and closed-loop reinforcement learning into a single infrastructure. Ultimately, these results demonstrate Droneulator as a highly practical pre-field integration layer that bridges high-fidelity visual rendering with realistic multirotor dynamics. Building upon this foundation, ongoing development will target multi-agent scalability, faster-than-real-time execution, and advanced environmental modelling to further enhance its utility, culminating in the public release of the open-source software stack.

\bibliographystyle{ieeetr}
\bibliography{references}

\end{document}